\documentclass{article}
\usepackage{ijcai24}

\usepackage{times}
\usepackage{soul}
\usepackage{url}
\usepackage[hidelinks]{hyperref}
\usepackage[utf8]{inputenc}
\usepackage[small]{caption}
\usepackage{graphicx}
\usepackage{amsmath}
\usepackage{amsthm}
\usepackage{booktabs}
\usepackage{algorithm}
\usepackage{algorithmic}
\usepackage[switch]{lineno}

\title{LangXAI: Integrating Large Vision Models for Generating Textual Explanations to Enhance Explainability in Visual Perception Tasks}

\author{
Hung Nguyen$^{1,3}$
\and
Tobias Clement$^2$\and
Loc Nguyen$^2$\and
Nils Kemmerzell$^2$\and\\
Binh Truong$^3$\and
Khang Nguyen$^3$\and
Mohamed Abdelaal$^4$\and
Hung Cao$^1$\\
\affiliations
$^1$Analytics Everywhere Lab, University of New Brunswick, Canada\\
$^2$Friedrich-Alexander-University Erlangen-Nürnberg, Germany\\
$^3$Quy Nhon AI, FPT Software, Vietnam\\
$^4$Software AG, Germany\\
\emails
hung.ntt@unb.ca, \{tobias.clement, loc.pt.nguyen, nils.kemmerzell\}@fau.de, \\\{binhtv8, khangnvt1\}@fpt.com,
mohamed.abdelaal@softwareag.com, hcao3@unb.ca
}

\begin{document}

\maketitle

\begin{abstract}
LangXAI is a framework that integrates Explainable Artificial Intelligence (XAI) with advanced vision models to generate textual explanations for visual recognition tasks. Despite XAI advancements, an understanding gap persists for end-users with limited domain knowledge in artificial intelligence and computer vision. LangXAI addresses this by furnishing text-based explanations for classification, object detection, and semantic segmentation model outputs to end-users. Preliminary results demonstrate LangXAI's enhanced plausibility, with high BERTScore across tasks, fostering a more transparent and reliable AI framework on vision tasks for end-users.
\end{abstract}
\section{Introduction}
In the field of artificial intelligence (AI), making complex AI decisions comprehensible is essential, especially in the application context requiring a large demand for explainability, such as healthcare, and banking. Explainable AI (XAI) is vital for achieving this, yet image-based XAI methods currently demand substantial AI and computer vision (CV) knowledge, often necessitating a domain expert to describe explanations to end-users~\cite{jin2019bridging,nguyen2023towards}. Large Vision Models (LVMs), evolving from Large Language Models (LLMs) for visual tasks, present a promising approach to address this issue. LVMs adeptly interpret visual data in human-like ways, improving AI system transparency. Recognizing the potential of XAI and LVM synergy, we aim to develop an innovative XAI framework that incorporates advanced LVMs to elucidate a broad spectrum of CV tasks comprehensively.

More specifically, we introduce LangXAI, a framework that employs LVMs to deliver clear text-based explanations for AI visual decision processes. LangXAI aims to increase the transparency of black-box models, making it easier for users without extensive specialized knowledge to understand AI decisions. Furthermore, during the validation stage, we employ a diverse set of metrics to thoroughly evaluate the model. Our initiative is dedicated to enhancing both AI transparency and trustworthiness through accessible explanations.

\begin{figure*}[tbh!]
    \centering
    \includegraphics[width=.85\textwidth]{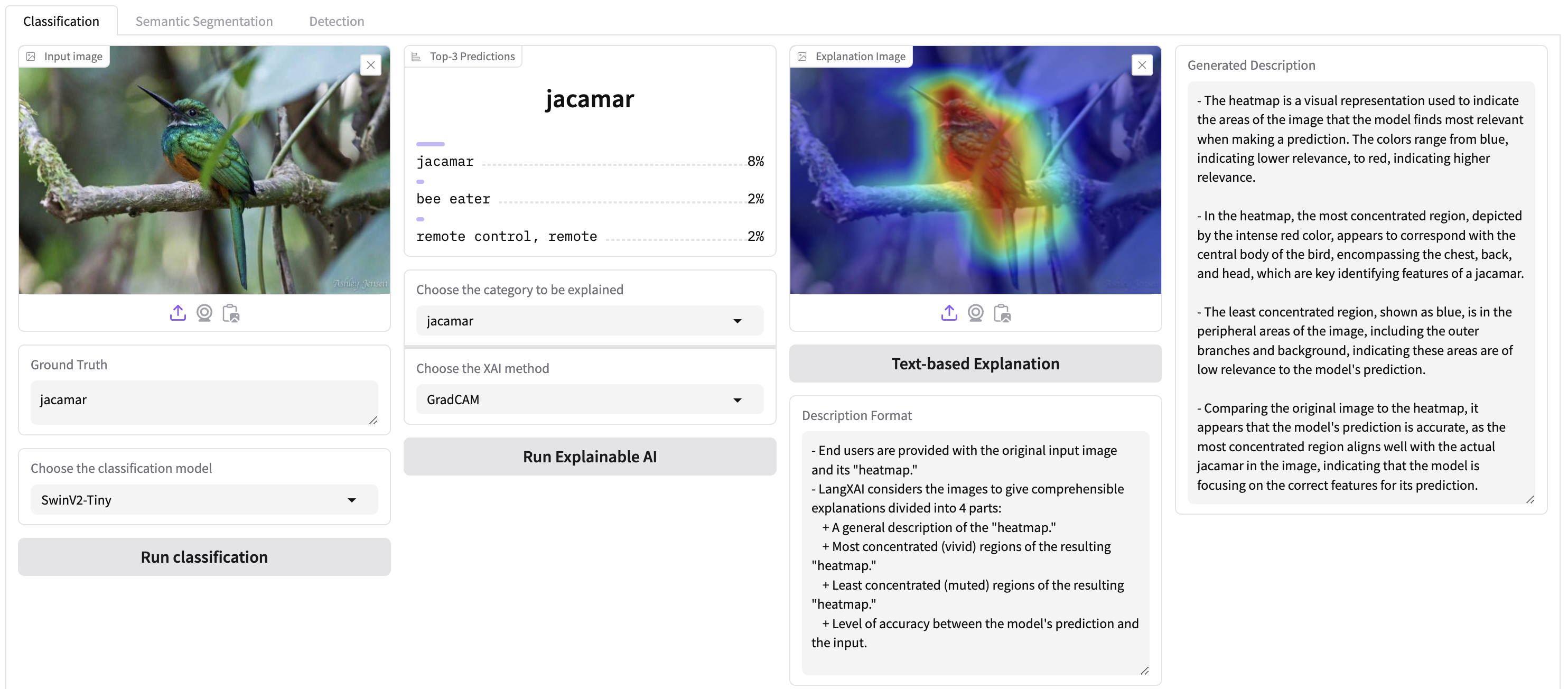}
    \caption{The interface of LangXAI showcases how it operates to make AI decisions in the classification task, which is designed straightforwardly with guidance so end-users can comprehend and monitor end-to-end explanations.}
    \label{fig:langxai_interface}
\end{figure*}

\section{Related Work}
We review two key aspects of our study. First, we investigate the current development of XAI in the fields of CV, with a specific focus on classification, semantic segmentation, and object detection. Additionally, we explore the emerging landscape of LVMs and their rapid development.

\subsection{Explainable AI (XAI) in CV}

\begin{figure}[htb!]
    \centering
    \includegraphics[width=.8\linewidth]{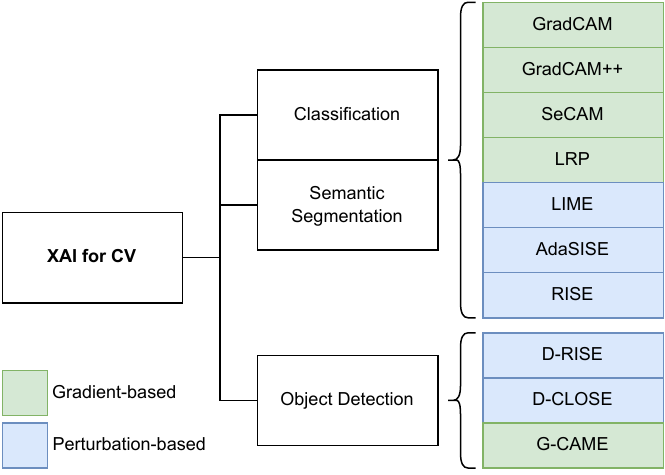}
    \caption{Classification of XAI methods by CV tasks and their mechanisms.}
    \label{fig:xai_clf}
\end{figure}

XAI methods for CV tasks can be categorized into two distinct groups based on their purposes, as illustrated in Figure~\ref{fig:xai_clf}. While classification and semantic segmentation share common XAI techniques due to their inherent similarities, object detection necessitates different approaches owing to its dual focus on classification and localization. In contrast to classification tasks that take into account the entire image, object detectors primarily employ convolutional layers instead of fully connected ones. This design choice confines the receptive field of the desired output to just a segment of the input image~\cite{truong2023towards}.

Within the context of working mechanisms, all XAI methods utilized in this study fall into two classes: Gradient-based and Perturbation-based. Gradient-based methods assess feature significance by computing gradients of the output concerning the extracted features using backpropagation. These methods then assign estimated attribution scores, identify the path that maximizes a specific output, and highlight critical input features, such as pixels in this study \cite{r2_1a}. We take into consideration several gradient-based algorithms, such as GradCAM \cite{r2_11b}, GradCAM++ \cite{r2_12}, SeCAM \cite{r2_1}, HiResCAM \cite{draelos2020use}, and G-CAME \cite{nguyen2023g}. 
In contrast, perturbation-based methods modify input images to track changes in the output. Significant output alternations indicate input relevance, especially when the target class is modified. These methods support iterative testing and offer visualizations of crucial input segments \cite{r2_1c}. In this research, several perturbation-based methods have been utilized, namely AdaSISE \cite{r2_1d}, RISE \cite{r2_1e}, D-RISE \cite{r2_40}, and D-CLOSE \cite{truong2023towards}.

\subsection{Large Vision Models (LVMs)}
The advancements in LLMs have given rise to LVMs, which blend language understanding and reasoning with visual perception \cite{r2_41}. Initial LVM strategies involve refining visual encoders based on language embeddings or visual-to-text conversion \cite{r2_42}.  Models such as Flamingo \cite{r2_43} and PaLM-E \cite{r2_44} exemplify the former approach, while techniques for the latter method have been proposed in \cite{r2_46} and \cite{r2_47}. In this research, we employed a new member of the LVMs family, which is GPT-4 Vision \cite{r2_49}. This LVM exhibits robust performance across diverse tasks \cite{r2_50} and demonstrates a strong alignment with human evaluators \cite{r2_51}. 

It is worth noting that, despite the presence of several end-user-centered XAI frameworks and patterns~\cite{jin2019bridging,jin2021euca,schoonderwoerd2021human}, contemporary research lacks an attempt to develop a unified framework for providing comprehensive and trustworthy end-to-end explanations to end-users for the saliency-map explanations \cite{r2_j,clement2024xai}. To address this gap, this paper aims to introduce a novel XAI framework that leverages the capabilities of LVMs to simplify user-system interaction and offer highly reliable interpretations.


\section{Framework}

\begin{figure*}[tbh!]
    \centering
    \includegraphics[width=\textwidth]{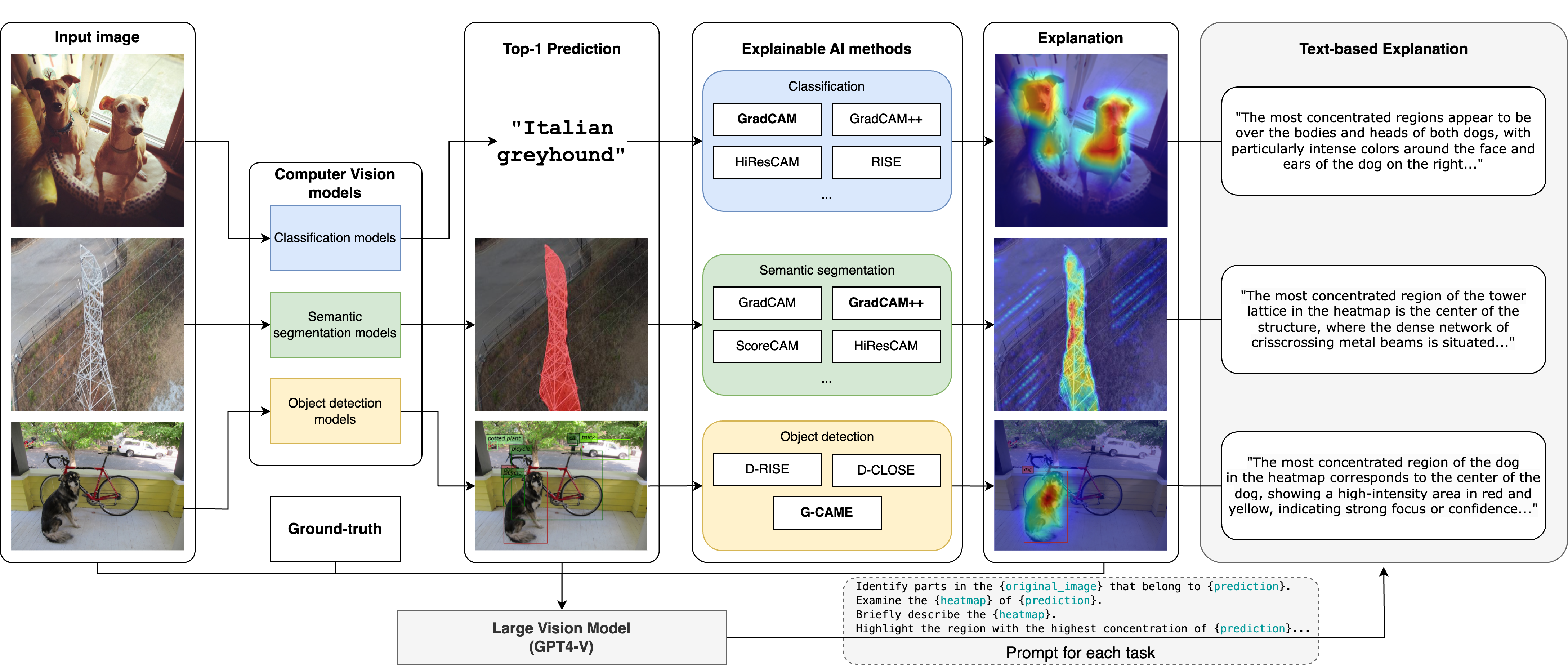}
    \caption{Our framework is split into two parts for explaining decisions made by AI models in CV tasks. The first part (in white blocks) generates saliency maps, where the XAI method in bold is used to generate the saliency map. The second part (in gray blocks) combines the input image, saliency map, ground truth, and prediction to provide a text-based explanation under prompts for each task. }
    \label{fig:langxai}
\end{figure*}


Our framework is divided into two main parts, aiming to provide end users with a straightforward approach to comprehending complex image decision explanations. The procedural details of the framework are depicted in Figure~\ref{fig:langxai}.

\subsection{Block 1: Saliency Map Extraction with XAI}
The first part of our framework focuses on generating saliency maps using XAI methods from various CV models tailored for different tasks. The process involves uploading an image and selecting the desired task, with specific models assigned accordingly: Swin Transformer v2~\cite{liu2022swin} for classification, DeepLabv3-ResNet50 and ResNet101~\cite{chen2017rethinking} for semantic segmentation, and Faster R-CNN~\cite{ren2015faster} and YOLOX~\cite{ge2021yolox} for object detection. Following the image analysis, users can specify the predicted class and choose the XAI method to generate saliency maps, which highlights the areas of interest for the model's decision-making process.

\subsection{Block 2: Text-based Explanation with LVM}
In the second part, we integrate various data to aid the LVM in generating text-based explanations for end-users unfamiliar with AI and CV. The GPT-4 Vision~\cite{r2_49} serves as the core LVM in our framework, leveraging information such as the input image, ground truth, model's top-1 prediction, and saliency map. We employ a structured prompt for each task, starting with presenting the image and saliency map to help the LVM identify focal areas. We then combine the saliency map with the model's prediction to verify the accuracy. In the end, we compare the model's prediction with the ground truth to determine the reliability and assess potential confusion by background or other objects. This comprehensive process ensures explanations are both coherent and indeed based on the model's visual analysis of the image.

\section{Evaluation}
In this section, we evaluate how well LangXAI's text-based explanations align with expert interpretations in the field of XAI. We ensure a comprehensive assessment by having a domain expert who has knowledge in AI and XAI context to review and label 5 samples for each task. The datasets chosen for analysis are tailored to each task: ImageNetv2~\cite{recht2019imagenet} for image classification, TTPLA~\cite{abdelfattah2020ttpla} for semantic segmentation, and MS-COCO 2017~\cite{cocodataset} for object detection. During the evaluation stage, we employ various metrics, including BLEU~\cite{papineni2002bleu}, METEOR~\cite{banerjee2005meteor}, ROUGE-L~\cite{lin2004rouge}, and BERTScore~\cite{zhang2019bertscore}, to comprehensively measure the performance of the LVM. Among these metrics, we prioritize BERTScore as it evaluates semantic similarity at a deeper level compared to surface lexical matches, providing a more nuanced reflection of expert interpretations. On the other hand, when assessing ROUGE-L and BERTScore, we particularly focus on their precision results.

\begin{table}[]
    \centering
    \resizebox{\linewidth}{!}{\begin{tabular}{ccccc}
    \toprule
       \textbf{Task}  &  BLEU & METEOR & ROUGE-L & BERTScore \\
    \midrule
       \textbf{Classification}  & 0.2971  & 0.5122  & 0.5196 & 0.9341 \\
       \textbf{Semantic Segmentation} & 0.2552 & 0.4741 & 0.4714 & 0.8594  \\
       \textbf{Object Detection} & 0.2754 & 0.4904 & 0.4911 & 0.9093 \\
   \bottomrule
    \end{tabular}
    }
    \caption{The performance of LVM across tasks is measured using BLEU, METEOR, ROUGE-L, and BERTScore metrics. These metrics evaluate how closely the model's text-based explanations align with expert interpretations in the context of XAI. Higher scores indicate better performance.}
    \label{tab:eval_table}
\end{table}
The evaluation results from Table~\ref{tab:eval_table} demonstrate varying performance levels of the LVM across different tasks: image classification, semantic segmentation, and object detection. It is worth noting that BERTScore metrics consistently yield high scores across all tasks, which indicates a robust semantic alignment between the model's explanations and expert interpretations. These results also demonstrate a deep comprehension of the model for the tasks at hand. Specifically, the image classification task receives the highest evaluation scores, likely owing to its straightforward nature of identifying depicted subjects in images. On the other hand, semantic segmentation and object detection demand more intricate explanations. They delve into object positions, the background, surrounding contextual information, and interactions within images, posing challenges in conveying them clearly to non-expert users.


\section{Conclusion}
This paper presents LangXAI, a novel framework combining XAI with advanced vision models to create textual explanations for visual tasks. Our framework can enhance purely visual explanations with natural language and therefore could help narrow the knowledge gap for users with limited AI expertise, highlighted by an average BERTScore of 0.9, indicating its potential to improve AI system transparency and reliability. 


\bibliographystyle{named}
\bibliography{ijcai24}

\end{document}